%% file: main_stega.tex
\documentclass[11pt,a4paper]{article}

\input{macro.tex}

\usepackage[hyperref]{emnlp2020}
\usepackage{times}
\usepackage{latexsym}

\usepackage{microtype}

\aclfinalcopy 
\def\aclpaperid{328} 

\definecolor{OrangeRed}{rgb}{1.0, 0.27, 0.0}

\title{Near-imperceptible Neural Linguistic Steganography \\ via Self-Adjusting Arithmetic Coding}

\author{Jiaming Shen, Heng Ji, Jiawei Han\\
\small Department of Computer Science, University of Illinois Urbana-Champaign, IL, USA \\
\footnotesize \{js2, hengji, hanj\}@illinois.edu \\
}

\date{}

\begin{document}
\maketitle

\begin{abstract}
\input{0-abstract}
\end{abstract}

\input{1-introduction.tex}

\input{2-problem.tex}
\input{3-methodology.tex}
\input{4-experiments.tex}

\input{5-related_work.tex}
\input{6-conclusion.tex}
\input{7-ack.tex}

\bibliographystyle{acl_natbib}
\bibliography{emnlp2020}

\end{document}

%% file: macro.tex
\usepackage{amsmath, amsfonts, amssymb, amsthm, xspace, color}
\usepackage{booktabs} 

\usepackage{subfigure, multirow, tabularx} 
\usepackage{booktabs} 
\usepackage{enumitem}
\usepackage{flushend}
\usepackage[T1]{fontenc}
\usepackage{epstopdf}
\usepackage{balance}
\usepackage{graphicx}
\usepackage[noend,ruled,linesnumbered]{algorithm2e} 
\usepackage{url}
\usepackage{wrapfig,lipsum}
\usepackage{makecell}
\usepackage{wrapfig}

\newcommand{\hide}[1]{} 

\newcommand{\etc}{\emph{etc.}\xspace} 
\newcommand{\ie}{i.e.\xspace} 
\newcommand{\eg}{e.g.\xspace} 
\newcommand{\nop}[1]{}

\newcommand{\naive}{na\"{\i}ve\xspace} 

\newtheorem{thm:def}{Definition}
\newtheorem{thm:eg}{Example}
\newtheorem{thm:lem}{Lemma}
\newtheorem{thm:obs}{Observation}
\newtheorem{thm:req}{Requirement}
\newtheorem{thm:prop}{Proposition}
\newtheorem{thm:principle}{Principle}
\newtheorem{thm:thm}{Theorem}
\newtheorem{thm:corollary}{Corollary}

\def \P {\mathbf{P}}

\newcommand{\our}{\mbox{\sc SAAC}\xspace}

\definecolor{midnightgreen}{rgb}{0.0, 0.29, 0.33}
\definecolor{orange}{RGB}{255,127,0}

\usepackage{bbm}
\usepackage{adjustbox}
\usepackage[super]{nth}

%% file: 0-abstract.tex

Linguistic steganography studies how to hide secret messages in natural language cover texts. 
Traditional methods aim to transform a secret message into an innocent text via lexical substitution or syntactical modification.
Recently, advances in neural language models (LMs) enable us to directly generate cover text conditioned on the secret message. 
In this study, we present a new linguistic steganography method which encodes secret messages using self-adjusting arithmetic coding based on a neural language model. 
We formally analyze the statistical imperceptibility of this method and empirically show it outperforms the previous state-of-the-art methods on four datasets by 15.3\% and 38.9\% in terms of bits/word and KL metrics, respectively. 
Finally, human evaluations show that 51\% of generated cover texts can indeed fool eavesdroppers.\footnote{\small Code and datasets are available at \url{https://github.com/mickeystroller/StegaText}.}

%% file: 1-introduction.tex
\section{Introduction}\label{sec:intro}
\vspace{-0.1cm}

Privacy is central to modern communication systems such as email services and online social networks.  
To protect privacy, two research fields are established: (1) \emph{cryptography} which encrypts secret messages into codes such that an eavesdropper is unable to decrypt, and (2) \emph{steganography} which encodes messages into cover signals such that an eavesdropper is not even aware a secret message exists~\cite{Westfeld1999AttacksOS, Amin2003InformationHU, Chang2014PracticalLS}. 
One useful cover signal for steganography is natural language text because of its prevalence and innocuity in daily life. 

Traditional linguistic steganography methods are mostly edit-based, i.e., they try to directly edit the secret message and transform it into an innocent text that will not raise the eavesdropper's suspicious eyes. 
Typical strategies include synonym substitution~\cite{Topkara2006TheHV}, paraphrase substitution~\cite{Chang2010LinguisticSU}, and syntactic transformation~\cite{Safaka2016MatryoshkaHS}, applied to various text media such as Email~\cite{Tutuncu2015NewAI} and Twitter~\cite{Wilson2014LinguisticSO}. 
Although being able to maintain the grammatical correctness of output text, those edit-based methods 
cannot encode information efficiently. 
For example, the popular CoverTweet system~\cite{Wilson2016AvoidingDO} can only encode two bits of information in each tweet on average. 

 \begin{figure}[!t]
   \centering
   \centerline{\includegraphics[width=0.48\textwidth]{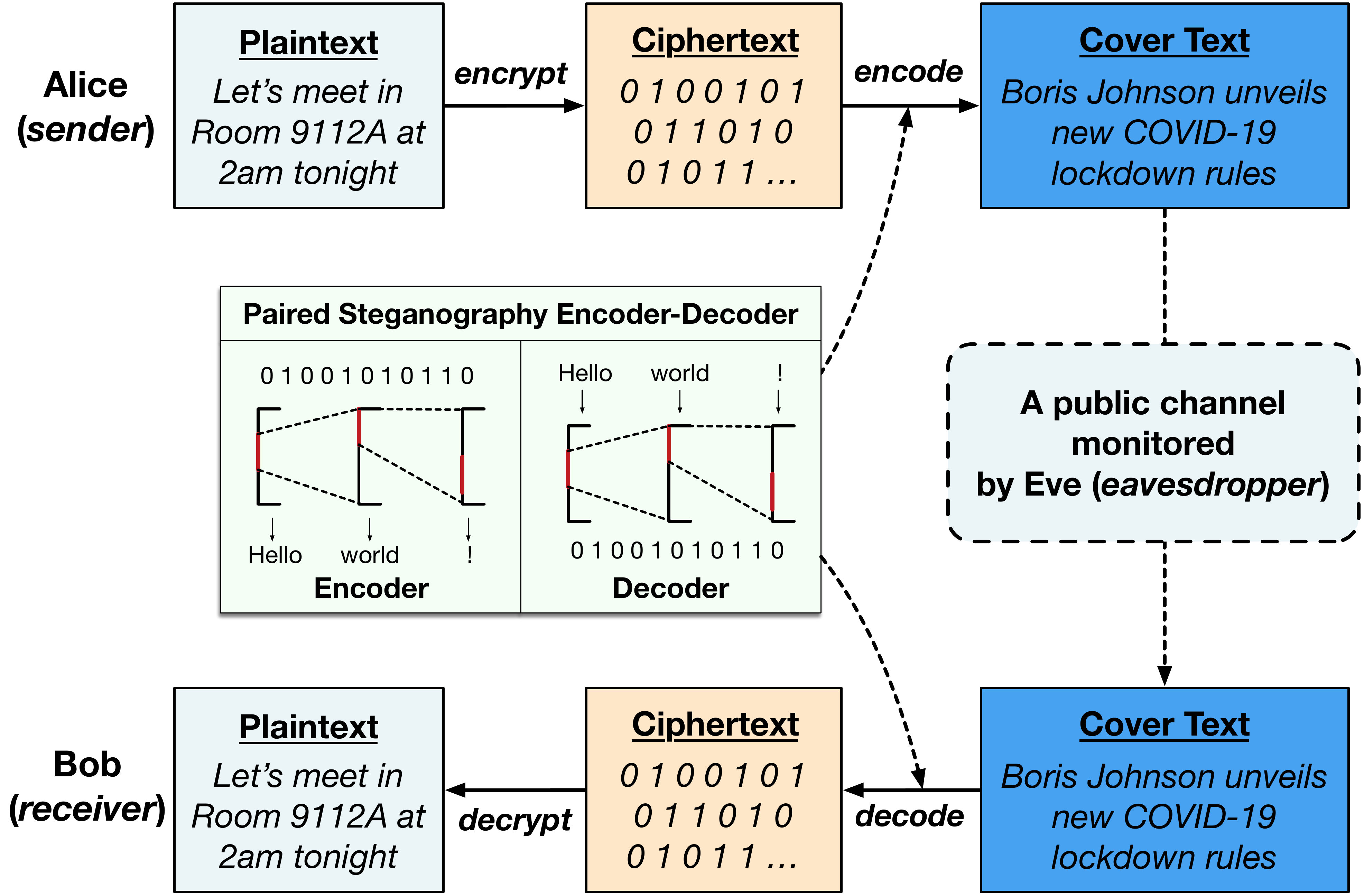}}
   \vspace{-0.2cm}
   \caption{Linguistic steganography pipeline.}
   \label{fig:pipeline}
   \vspace{-0.3cm}
 \end{figure}
 
Recent advances in neural language models (LMs)~\cite{Jzefowicz2016ExploringTL, radford2019language, Yang2019XLNetGA} have enabled a diagram shift from edit-based methods to generation-based methods which directly output a cover text by encoding the message reversibly in the choices of tokens. 
Various encoding algorithms~\cite{Fang2017GeneratingST, Yang2019RNNStegaLS, Ziegler2019NeuralLS} have been proposed to leverage neural LMs to generate high-quality cover texts in terms of both fluency and information hiding capacity. 
However, most of the existing methods do not provide explicit guarantees on the imperceptibility of generated cover text (\ie, to what extent the cover text is indistinguishable from natural texts without hidden messages). 
One recent exception is the work~\cite{Dai2019TowardsNS} which shows the imperceptibility of the method in~\citet{Fang2017GeneratingST}. 
Nevertheless, for other more advanced steganography methods~\cite{Yang2019RNNStegaLS, Ziegler2019NeuralLS}, their imperceptibilities still remain unknown. 
 
In this work, we propose a new linguistic steganography method with guaranteed imperceptibility. 
Our new method is built based on the previous study~\cite{Ziegler2019NeuralLS} which views each secret message as a binary fractional number and encodes it using arithmetic coding~\cite{Rissanen1979ArithmeticC} with a pretrained neural LM. 
This method generates cover text tokens one at a time (c.f. Fig.~\ref{fig:arithmetic}).
At each time step $t$, it computes a distribution of the $t$-th token using the given LM; truncates this distribution to include only top $K$ most likely tokens, and finally outputs the $t$-th token based on the secret message and the truncated distribution.
In their study, this hyperparameter $K$ is the same across all generation steps. 
We analyze this method's imperceptibility and show it is closely related to the selected $K$. 
Specifically, increasing $K$ will improve the method's imperceptibility at the cost of a larger probability of generating rarely-used tokens and slower encoding speed. 
When the cover text token distribution is flat and close to the uniform distribution, we need a large $K$ to achieve the required imperceptibility guarantee.
When the cover text token distribution is concentrated, we can use a small $K$ to avoid generating rarely-used tokens and to increase encoding speed.
As different generation steps will witness different underlying cover text token distributions, using a static $K$ is clearly sub-optimal. 

To address this issue, we propose a new algorithm \our\footnote{\small SAAC is short for \underline{S}elf-\underline{A}djusting \underline{A}rithmetic \underline{C}oding.} which automatically adjusts $K$ by comparing the truncated cover text token distribution with the original LM's output at each generation step and selects the minimal $K$ that achieves the required imperceptibility. 
We theoretically prove the \our algorithm is near-imperceptible for linguistic steganography and empirically demonstrate its effectiveness on four datasets from various domains.
Furthermore, we conduct human evaluations via crowdsourcing and show 51\% of cover texts generated by \our can indeed fool eavesdropper.

\smallskip
\noindent \textbf{Contributions.}~
This study makes the following contributions: 
(1) We formally analyze the imperceptibility of arithmetic coding based steganography algorithms;
(2) We propose \our, a new near-imperceptible linguistic steganography method that encodes secret messages using self-adjusting arithmetic coding with a neural LM; and
(3) Extensive experiments on four datasets demonstrate our approach can on average outperform the previous state-of-the-art method by 15.3\% and 38.9\% in terms of bits/word and KL metrics, respectively.

%% file: 2-problem.tex
\section{Background}\label{sec:problem}

\subsection{Linguistic Steganography}
We consider the following scenario where Alice (sender) wants to send Bob (receiver) a secret message (plaintext) through a public text channel (\eg, Twitter and Reddit) monitored by Eve (eavesdropper).
This is also known as the ``prisoner problem''~\cite{simmons1984prisoners}.  
Eve expects to see fluent texts in this public channel and will suspect every non-fluent text of concealing some hidden messages. 
Therefore, Alice's goal is to transform the plaintext into a fluent cover text that can pass through Eve's suspicious eyes while ensuring that only Bob can read the secret message.

To achieve this goal, Alice could take the ``encrypt-encode'' approach (c.f. Fig.~\ref{fig:pipeline}).
Namely, she first encrypts the plaintext into a ciphertext (\ie, a bit sequence indistinguishable from a series of fair coin flips) and then encodes the ciphertext into the cover text using an encoder $f$. 
When Bob receives the cover text, he first decodes it into the ciphertext using the decoder $f^{-1}$ and then decrypts the ciphertext into the plaintext. 
Linguistic steganography research focuses on the encoding/decoding steps, \ie, how to design the encoder that transforms the bit sequence into a fluent cover text and its paired decoder that maps the cover text back to the original bit sequence. 
Note here we introduce the middle ciphertext for two purposes.
First, it increases communication security as more advanced encryption/decryption methods (\eg, AES, RSA, \etc) can be used on top of the steganography encoder/decoder. 
Second, it enlarges the output cover text space by removing the unnecessary restriction that the cover text must be transformed from the original plaintext.

\subsection{Statistical Imperceptibility}\label{subsec:stats_imp}
\smallskip
\noindent \textbf{Notations.} 
A vocabulary $\mathcal{V}$ is a finite set of tokens\footnote{\small Each token can be a single word, a subword unit, or even a character, depending on the tokenizer choice.}.
A language model (LM) inputs a token sequence $\mathbf{x} = [x_1, x_2, \dots, x_n]$ and returns the joint probability $\P_{LM}(\mathbf{x})$. 
From this joint probability, we can derive the conditional probability $\P_{LM}(x_{t+1}|x_1, \dots, x_{t})$ which enables us to sample a text $\mathbf{x}$ by drawing each token $x_{t}, t = 1, 2, \dots$, one at a time.

A steganography encoder $f$ inputs a language model $\P_{LM}$ as well as a length-$L$ ciphertext $\mathbf{m} \sim \text{Unif}(\{0, 1\}^{L})$, and outputs its corresponding cover text $\mathbf{y} = f(\mathbf{m};\P_{LM})$. 
To ensure the receiver can uniquely decode the cover text, this encoder function $f$ must be both deterministic and invertible. 
Moreover, this encoder $f$, together with the ciphertext distribution and the input LM, implicitly define a distribution of cover text $\mathbf{y}$ which we denote as $\mathbf{Q}(\mathbf{y})$.
When cover texts are transmitted in the public channel, this distribution $\mathbf{Q}(\mathbf{y})$ is what an eavesdropper would observe. 

\smallskip
\noindent \textbf{Imperceptibility.}\label{subsec:imp}
To avoid raising eavesdropper's suspicion, we want the cover text distribution $\mathbf{Q}$ to be similar to the true natural language distribution (\ie, what this eavesdropper would expect to see in this public channel). 
Following~\cite{Dai2019TowardsNS}, we formulate ``imperceptibility'' using the total variation distance (TVD) as follows:
\begin{equation}
\small
\text{TVD}(\mathbf{P}^{*}_{LM}, \mathbf{Q}) = \frac{1}{2} \|\mathbf{Q} - \mathbf{P}^{*}_{LM}\|_{1},
\end{equation}
where $\mathbf{P}^{*}_{LM}$ denotes the true language distribution. 
As we approximate $\mathbf{P}^{*}_{LM}$ using a LM $\mathbf{P}_{LM}$ (\eg, OpenAI GPT-2~\cite{radford2019language}), we further decompose $\text{TVD}(\mathbf{P}^{*}_{LM}, \mathbf{Q})$ as follows:
\begin{equation}
\small
\text{TVD}(\mathbf{P}^{*}_{LM}, \mathbf{Q}) \leq \frac{1}{2} \|\mathbf{P}^{*}_{LM} - \mathbf{P}_{LM}\|_{1} + \frac{1}{2} \|\mathbf{P}_{LM} - \mathbf{Q}\|_{1},
\end{equation}
where the first term measures how good this LM is and the second term, that is the main focus of this study, indicates the gap induced by the steganography encoder. 
Even without knowing the first term, we can still obtain a relative imperceptibility guarantee based on the second term, which enables us to compare different steganography algorithms. 

Using Pinsker's inequality~\cite{Fedotov2003RefinementsOP}, we set the upper-bound for the total variation distance using the KL divergence\footnote{\small We will consistently compute KL divergence in base 2.}:
\begin{equation}
\small
\frac{1}{2} \|\mathbf{P}_{LM} - \mathbf{Q} \|_{1}  \leq \sqrt{\frac{\text{ln}2}{2} D_{KL} (\mathbf{Q} \| \mathbf{P}_{LM})}.
\end{equation}
Then, we further decompose the right hand side of the above inequality based on the additivity of KL divergence and obtain the following result:
\begin{equation}\label{eq:decompose}
\small
\frac{1}{2} \|\mathbf{P}_{LM} - \mathbf{Q} \|_{1}  \leq \sqrt{\frac{\text{ln}2}{2} \sum_{t=1}^{\infty} D_{KL} (\mathbf{Q}(\cdot | \mathbf{y}_{<t}) \| \mathbf{P}_{LM}(\cdot | \mathbf{y}_{<t}))},
\end{equation}
where $\mathbf{y}_{<t}=[y_1, \dots, y_{t-1}]$ is a cover text prefix. 
$\mathbf{P}_{LM}(\cdot | \mathbf{y}_{<t})$ and $\mathbf{Q}(\cdot | \mathbf{y}_{<t})$ are distributions over the next token $y_t$ conditioned on the prefix $\mathbf{y}_{<t}$ before and after the steganography encoding algorithm, respectively. 
This inequality provides a formal framework to analyze the imperceptibility of a steganography encoder.
Moreover, it implies that in order to achieve the near-imperceptibility, we must guarantee the encoder's output $\mathbf{Q}(\cdot | \mathbf{y}_{<t})$ being close to its input $\mathbf{P}_{LM}(\cdot | \mathbf{y}_{<t})$ at all steps.

%% file: 3-methodology.tex
\section{Self-Adjusting Arithmetic Coding}\label{sec:method}

In this section, we first introduce the general arithmetic coding and discuss its practical limitations.
We then present \our, a self-adjusting arithmetic coding algorithm and analyze its imperceptibility.

\subsection{Arithmetic Coding}

Arithmetic coding is a method initially proposed to compress a string of elements sampled from a known probability distribution~\cite{Rissanen1979ArithmeticC}. 
For data compression, arithmetic coding is asymptotically optimal in the sense that it can compress information within a long string to its entropy. 
In practice, it also outperforms the better-known Huffman coding method~\cite{huffman1952method} because it does not partition the input string into blocks. 
Traditionally, arithmetic coding encodes a string of elements into a bit sequence. 
To use such a coding for linguistic steganography, we follow~\cite{Ziegler2019NeuralLS} and reverse the encoding order. 
Namely, we \emph{encode} a bit sequence (ciphertext) into a string of tokens (cover text) and \emph{decode} a cover text to its original ciphertext.

 \begin{figure*}[!t]
   \centering
   \centerline{\includegraphics[width=0.92\textwidth]{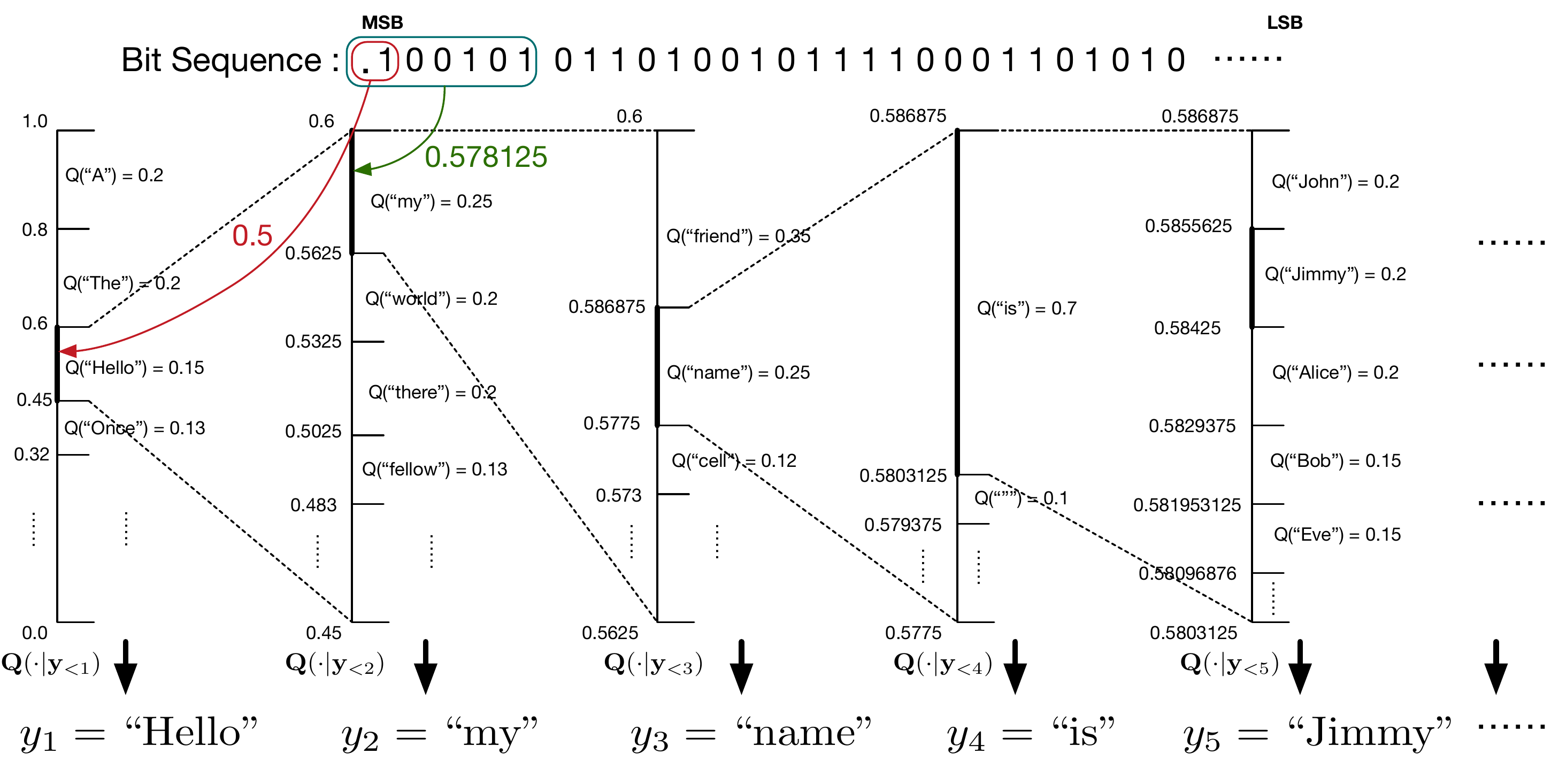}}
   \vspace{-0.2cm}
   \caption{A running example of arithmetic coding. We input a bit sequence (\ie, the ciphertext) with the most significant bit (MSB) at the left and output the encoded cover text.}
   \label{fig:arithmetic}
   \vspace{-0.3cm}
 \end{figure*}
 
\smallskip
\noindent \textbf{Encoding.}
During the encoding stage, we view the bit sequence $\mathbf{m} = [m_1, m_2, \dots, m_L]$ as the binary representation of a single number $B(\mathbf{m}) = \sum_{i=1}^{L} m_{i} \times 2^{-i}$. 
For example, if $\mathbf{m} = [1, 0, 1]$, we have $B(\mathbf{m}) = 1 \times 2^{-1} + 1 \times 2^{-3} = 0.625$. 

The encoder generates the cover text token one at a time.
At each time step $t$, the encoder has access to an underlying language model $\P_{LM}$ and considers three things: (1) the number $B(\mathbf{m})$, (2) the cover text prefix $\mathbf{y}_{<t}$, and (3) the current interval $[l_{t}, u_{t})$ (at the beginning of the encoding process, this interval $[l_1, u_1)$ is set to [0, 1), but it will change). 
Based on the LM and cover text prefix, the encoder first computes the conditional distribution of the next token $\mathbf{Q}(y_t|\mathbf{y}_{<t})$. 
Then, it divides the current interval $[l_{t}, u_{t})$ into sub-intervals, each representing a fraction of the current interval proportional to the conditional probability of a possible next token. 
Whichever interval contains the number $B(\mathbf{m})$ becomes the interval used in the next step (\ie, $[l_{t+1}, u_{t+1})$) and its corresponding token becomes the cover text token $y_t$. 
The encoding process stops when all $\mathbf{m}$-prefixed fractions fall into the final interval, that is, the generated cover text unambiguously defines the bit sequence $\mathbf{m}$. 
Before we discuss and analyze the concrete design of $\mathbf{Q}(\cdot|\mathbf{y}_{<t})$ in the next section, we first present a running example in Figure~\ref{fig:arithmetic}. 

Suppose we want to encode a bit sequence $\mathbf{m} = [1, 0, 0, 1, 0, 1, \dots]$. 
This bit sequence represents a fraction $B(\mathbf{m}) \in [0.58425, 0.58556)$. 
At the time step $t=1$, we divide the initial interval $[0, 1)$ and find $B(\mathbf{m})$ falling into the sub-interval $[0.45, 0.6)$ which induces the first cover text token $y_1 = \text{``Hello''}$.
At the time step $t=2$, we further divide the interval $[0.45, 0.6)$ and observe that $B(\mathbf{m})$ belongs to the range $[0.5625, 0.6)$ corresponding to the second cover text token $y_2 = \text{``my''}$.
We repeat this process until the final interval covers all binary fractions starting with $\mathbf{m}$ and output the generated cover text by then. 

\smallskip
\noindent \textbf{Decoding.}
During the decoding stage, we are given a cover text $\mathbf{y} = [y_1, \dots, y_n]$ as well as the same language model $\P_{LM}$ used in the encoding stage, and aim to recover the original ciphertext $\mathbf{m}$. 
We achieve this goal by reversing the encoding process and gradually narrowing the range of possible bit sequences. 
At each time step $t$, the decoder first generates the conditional distribution $\mathbf{Q}(y_t|\mathbf{y}_{<t})$.
Then, it divides the current interval $[l_{t}, u_{t})$ (initialized to $[0, 1)$) into sub-intervals based on $\mathbf{Q}(y_t|\mathbf{y}_{<t})$ and the one corresponding to $y_{t}$ becomes the interval used in the next step, that is, $[l_{t+1}, u_{t+1})$. 
The decoding process stops after we process the last cover text token $y_n$ and outputs the decoded ciphertext to be the shared common prefix of the binary representations of $l_{n+1}$ and $u_{n+1}$. 

\subsection{Imperceptibility Analysis}

One important issue remained in the general arithmetic coding procedure is how to design the conditional distribution $\mathbf{Q}(\cdot|\mathbf{y}_{<t})$. 
As we discussed in Section~\ref{subsec:imp}, this distribution should be close to the underlying model LM. 
Ideally, we may just set $\mathbf{Q}(\cdot|\mathbf{y}_{<t})$ to be the same as $\P_{LM}(\cdot|\mathbf{y}_{<t})$.
However, this \naive design has several problems.
First, it may generate a rarely-used cover text token because we are actually reading off the tokens based on the ciphertext, instead of really sampling the LM. 
This could harm the cover text fluency and raises the eavesdropper's suspicion.  
Second, $\P_{LM}(\cdot|\mathbf{y}_{<t})$ is a distribution over the entire vocabulary $\mathcal{V}$ (with a full rank $|\mathcal{V}|$) and using it to divide the $[0, 1)$ interval will quickly encounter the precision problem, even if we implement the coding scheme using a fixed precision binary fractions~\cite{Witten1987ArithmeticCF}. 
Finally, this design further slows down the coding speed and the slow speed is the major weakness of arithmetic coding compared to its rival Huffman method~\cite{Duda2013AsymmetricNS}. 

Due to the above reasons, people in practice will truncate the LM distribution to include only top $K$ most likely tokens~\cite{Ziegler2019NeuralLS}, which leads to the following distribution:
\begin{equation}
\small
\mathbf{Q}(y_{t} | \mathbf{y}_{<t}) \propto \left\{  \begin{array}{ll}
 \mathbf{P}_{LM}(y_{t}|\mathbf{y}_{<t}) & \text{if} ~~ y_{t} \in \mathcal{T}_{K}(\mathbf{y}_{<t}) \\
 0 & \text{otherwise} 
 \end{array}
 \right.
\end{equation}
where $\mathcal{T}_{K}(\mathbf{y}_{<t}) = \texttt{argtopK}_{y'}~\mathbf{P}_{LM}(y'|\mathbf{y}_{<t})$. 
Accordingly, we have the imperceptibility of one generation step to be:
\begin{align}\label{eq:dkl_original}
	& \small D_{KL}(\mathbf{Q}(y_t|\mathbf{y}_{<t}) \| \mathbf{P}_{LM}(y_t|\mathbf{y}_{<t})) = -\log Z_{K},  \nonumber \\ 
	& \small Z_{K} = \sum_{y' \in \mathcal{T}_{K}(\mathbf{y}_{<t})} \mathbf{P}_{LM}(y' | \mathbf{y}_{<t}),
\end{align}
where $Z_{K}$ is essentially the cumulative probability of top $K$ most likely tokens. 
From this equation, we can see that the imperceptibility of arithmetic coding depends crucially on how the underlying LM distribution concentrates on its top $K$ predictions.
Previous study uses the same $K$ across \emph{all} generation steps and ignores the different distribution characteristics in different steps.
This strategy is sub-optimal because in some steps, the pre-defined $K$ is too small to achieve good imperceptibility, while in the other steps, the same $K$ is too large and slows down the encoding speed.

In this study, we propose a new \emph{self-adjusting} arithmetic coding algorithm \our to remedy the above problem. 
The idea is to dynamically select the most appropriate $K$ that satisfies a pre-defined per-step imperceptibility guarantee. 
Specifically, the sender can set a small per-step imperceptibility gap $\delta \ll 1$ and at time step $t$, we set the $K_{t}$ as:
\begin{equation}
\small
K_t = \text{min} (\{K| \sum_{y' \in \mathcal{T}_{K}(\mathbf{y}_{<t})} \mathbf{P}_{LM}(y' | \mathbf{y}_{<t}) \geq  2^{-\delta} \} ).
\end{equation}
This selected $K_t$ is essentially the smallest $K$ that can achieve the imperceptibility guarantee.  
As we later show in the experiment, this selected $K$ varies a lot in different steps, which further confirms the sub-optimality of using a static $K$. 

The above method guarantees that each step incurs no more additional imperceptibility than the threshold $\delta$.
This makes the imperceptibility of \emph{an entire sequence} dependent on the length of bit sequence. 
To achieve a length-agnostic imperceptibility bound, we may choose a convergent series for per-step threshold. For example, if we set $\delta_{t} = \frac{\delta_{0}}{t^2}$ and based on the inequality~\ref{eq:decompose} we will have:
\begin{equation}
\small
\frac{1}{2} \|\mathbf{P}_{LM} - \mathbf{Q} \|_{1}  \leq \sqrt{\frac{\text{ln}2}{2} \sum_{t=1}^{\infty} \frac{\delta_{0}}{t^2} } = \sqrt{\frac{\pi^2\text{ln}2}{12} \delta_{0}}. 
\end{equation}
This result shows our proposed \our algorithm is near-imperceptible for linguistic steganography. 

%% file: 4-experiments.tex
\section{Experiments}\label{sec:exp}

    \begin{table}[t]
    \centering
    \scalebox{0.82}{
        \small
        \begin{tabular}{c|cccc}
            \toprule
            \textbf{Dataset} & \textbf{Drug} & \textbf{News} & \textbf{COVID-19} & \textbf{Random} \\
            \midrule
            Num. of Sentences & 3972 & 6437 & 2000 & 3000 \\
            Avg. Num. of Words & 19.01 & 14.30 & 24.21 & --- \\
            Avg. Num. of Bits & 289.75 & 211.08 & 308.65 & 256 \\ 
            \bottomrule
        \end{tabular}
    }
    \vspace{-0.2cm}
    \caption{Datasets statistics.}\label{table:dataset}
    \vspace{-0.3cm}
    \end{table}
    
    \begin{table*}[!t]
    \centering
    \scalebox{0.75}{
    \begin{tabular}{c|cccccccc}
    \toprule
    \multirow{2}{*}{\textbf{Methods}}   & \multicolumn{2}{c}{\textbf{Drug}}      & \multicolumn{2}{c}{\textbf{News}}      & \multicolumn{2}{c}{\textbf{COVID-19}}  & \multicolumn{2}{c}{\textbf{Random}}                    \\
    \cmidrule[0.06em](r){2-3} \cmidrule[0.06em](r){4-5} \cmidrule[0.06em](r){6-7} \cmidrule[0.06em](r){8-9}
                               & \emph{Bits/Word} $\uparrow$         & $D_{KL}$   $\downarrow$      & \emph{Bits/Word}  $\uparrow$        & $D_{KL}$   $\downarrow$  & \emph{Bits/Word}   $\uparrow$       & $D_{KL}$  $\downarrow$    & \emph{Bits/Word}   $\uparrow$       & $D_{KL}$  $\downarrow$ \\ 
                               
    \midrule
    Bin-LM ($B=1$)                   & 1          & 1.864          & 1          & 1.922          & 1          & 1.838   & 1          &  1.185         \\ 
    Bin-LM ($B=2$)                   & 2          & 2.358          & 2          & 2.385          & 2          & 2.346   & 2          & 2.374         \\ 
    Bin-LM ($B=3$)                   & 3          & 2.660          & 3          & 2.680          & 3          & 2.659   & 3          & 2.664         \\ 
    \midrule
    RNN-Stega ($H=3$)         & 2.370          &  1.015         & 2.387          & 1.015          & 2.368          & 0.999   & 2.378          & 0.991         \\ 
    RNN-Stega ($H=5$)         & 3.399          & 0.628          & 3.393          & 0.628          & 3.368           & 0.624   & 3.370          & 0.630         \\ 
    RNN-Stega ($H=7$)         & 4.202          & 0.424          & 4.202          & 0.426          & 4.197          & 0.426   & 4.163          & 0.422         \\ 
    \midrule
    Patient-Huffman ($\epsilon=0.8$)         & 1.835         &  0.269         &  1.834         &  0.269          & 1.844          & 0.270   & 1.847          & 0.271        \\ 
    Patient-Huffman ($\epsilon=1.0$)         & 2.147          & 0.360          & 2.154          & 0.361          & 2.142          & 0.357   & 2.148          & 0.358         \\ 
    Patient-Huffman ($\epsilon=1.5$)         & 2.596          & 0.524          & 2.583         & 0.522          & 2.579           & 0.519   & 2.584          & 0.520         \\ 
    \midrule
    Arithmetic ($K=300$)                & 3.497          & 0.203          & 3.491          &  0.209         & 3.510         & 0.191  & 3.466          & 0.189         \\ 
    Arithmetic ($K=600$)                & 4.247          & 0.162          & 4.240          & 0.166          & 4.289          & 0.146   & 3.599          & 0.160         \\ 
    Arithmetic ($K=900$)                & 4.376          & 0.149          & 4.358          & 0.152          & 4.414          & 0.131   & 3.669          & 0.147         \\ 
    \midrule
    \our ($\delta=0.1$)   & 4.262          & 0.153          & 4.232          & 0.157          & 4.301          & 0.133                 & 4.225          & 0.136         \\ 
    \our ($\delta=0.05$) & 4.451          & 0.134          & 4.441          & 0.138          & 4.519          & 0.114            & 4.419         & 0.117         \\ 
    \our ($\delta=0.01$) & \textbf{4.862}          & \textbf{0.109}          & \textbf{4.784}          & \textbf{0.117}          & \textbf{4.851}          & \textbf{0.093}   & \textbf{4.778}          & \textbf{0.099}         \\ 
    \bottomrule
    \end{tabular}
    }
    \vspace*{-0.2cm}
    \caption{\label{table:results} Quantitative performance of linguistic steganography methods across all datasets. Each method has one parameter controlling various tradeoffs (c.f. detailed discussions in Compared Method subsection) and we indicate them in the parentheses. Larger \emph{bits/word} values $\uparrow$ and smaller $D_{KL}$ values $\downarrow$ indicate better performance.}
    \vspace*{-0.3cm}
    \end{table*}
    
\subsection{Experiment Setups}

\noindent \textbf{Datasets.}
We conduct our experiments on four datasets from different domains: 
(1) \textbf{Drug}~\cite{Ji2018CreativeLE}, which contains a set of Reddit comments related to drugs, 
(2) \textbf{News}, which includes a subset of news articles in the CNN/DailyMail dataset~\cite{Hermann2015TeachingMT},
(3) \textbf{COVID-19}, which is a subset of research papers related to COVID-19 in the CORD-19 dataset~\cite{Wang2020CORD19TC}, and
(4) \textbf{Random}, which is a collection of uniformly sampled bit sequences. 
The first three datasets contain natural language texts and we convert them into bit sequences\footnote{\small This is essentially the encryption step in Fig.~\ref{fig:pipeline}.} following the same process in~\citet{Ziegler2019NeuralLS}.
Table~\ref{table:dataset} summarizes the dataset statistics. 
    
\smallskip
\noindent \textbf{Compared Methods.}
We compare the following linguistic steganography methods.
\begin{enumerate}[leftmargin=*,nosep]
        \item Bin-LM~\cite{Fang2017GeneratingST}: This method first splits the vocabulary $\mathcal{V}$ into $2^{B}$ bins and represents each bin using a $B$-bit sequence. Then, it chunks the ciphertext into $\lceil L/B \rceil$ blocks and encodes the $t$-th block by taking the most likely token (determined by the underlying LM) that falls in the $t$-th bin.
        \item RNN-Stega~\cite{Yang2019RNNStegaLS}: This method first constructs a Huffman tree for top $2^{H}$ most likely tokens at each time step $t$ according to $\P_{LM}(\cdot|\mathbf{y}_{<t})$. Then, it follows the bits in ciphertext to sample a cover text token $y_t$ from the constructed Huffman tree. It improves the above Bin-LM method by encoding one or more bits per generated cover text token.  
        \item Patient-Huffman~\cite{Dai2019TowardsNS}: This method improves RNN-Stega by explicitly checking if the KL divergence between the LM distribution and the Huffman distribution is smaller than a specified threshold $\epsilon$. If the KL divergence is larger than $\epsilon$, it samples from the base LM distribution and \emph{patiently} waits for another opportunity. 
        \item Arithmetic~\cite{Ziegler2019NeuralLS}: This method also uses the arithmetic coding to generate cover text tokens. At each time step $t$, it truncates the $\P_{LM}(\cdot|\mathbf{y}_{<t})$ distribution to include only top $K$ most likely tokens and samples one cover text tokens from the truncated distribution.
        \item \our: This method is our proposed \underline{S}elf-\underline{A}djusting \underline{A}rithmetic \underline{C}oding algorithm which automatically adjusts $\P_{LM}(\cdot|\mathbf{y}_{<t})$ to achieve the required imperceptibility guarantee $\delta$. 
\end{enumerate}

\smallskip
\noindent \textbf{Evaluation Metrics.}
We follow previous studies and evaluate the results using two metrics: 
    \begin{enumerate}[leftmargin=*,nosep]
        \item \emph{Bits/word}: This metric is the average number of bits that one cover text token can encode. A larger bits/word value indicates the algorithm can encode information more efficiently. 
        \item $D_{KL}$: This metric is the KL divergence between the LM distribution and the cover text distribution. A smaller $D_{KL}$ value indicates the model has better imperceptibility (c.f. Section~\ref{subsec:stats_imp}). 
\end{enumerate}

\smallskip
\noindent \textbf{Implementation Details.}
We implement all compared methods based on the codebase in~\cite{Ziegler2019NeuralLS}.
All the code and data are publicly available\footnote{\small \url{https://github.com/mickeystroller/StegaText}}.
Specifically, we use PyTorch 1.4.0 and the pretrained OpenAI GPT-2 medium model in the Huggingface library as the underlying LM for all methods.
This LM includes 345M parameters and there is no additional parameter introduced by steganography encoding algorithms.
For baseline method Bin-LM, we choose its block size $B$ in $[1,2,3,4,5]$.
For RNN-Stega method, we vary the Huffman tree depth $H$ in $[3,5,7,9,11]$.
For Patient-Huffman method, we change the patience threshold $\epsilon$ in $[0.8, 1.0, 1.5]$. 
For Arithmetic method, we select its hyperparameter $K$ ranging from 100 to 1800 with an increment 300 and fix its temperature parameter $\tau=1$. 
Finally, we choose the imperceptibility gap $\delta$ in our \our method in $[0.01, 0.05, 0.1]$. 
For both Arithmetic and \our methods, we implement the arithmetic coding using a fixed 26-bits precision binary fractions. 
We do not perform any hyperparameter search and directly report all the results in the main text.

\smallskip
\noindent \textbf{Discussions on LM Sharing.}
We note that all compared methods require the employed LM to be shared between the sender and the receiver beforehand.
Therefore, in practice, people typically use a popular public language model (e.g., GPT2) available to everyone. 
This allows two parties to directly download the same LM from a centroid place (e.g., an OpenAI hosted server) and removes the necessity of sending the LM though some communication channel.

 \begin{figure}[!t]
   \centering
   \centerline{\includegraphics[width=0.46\textwidth]{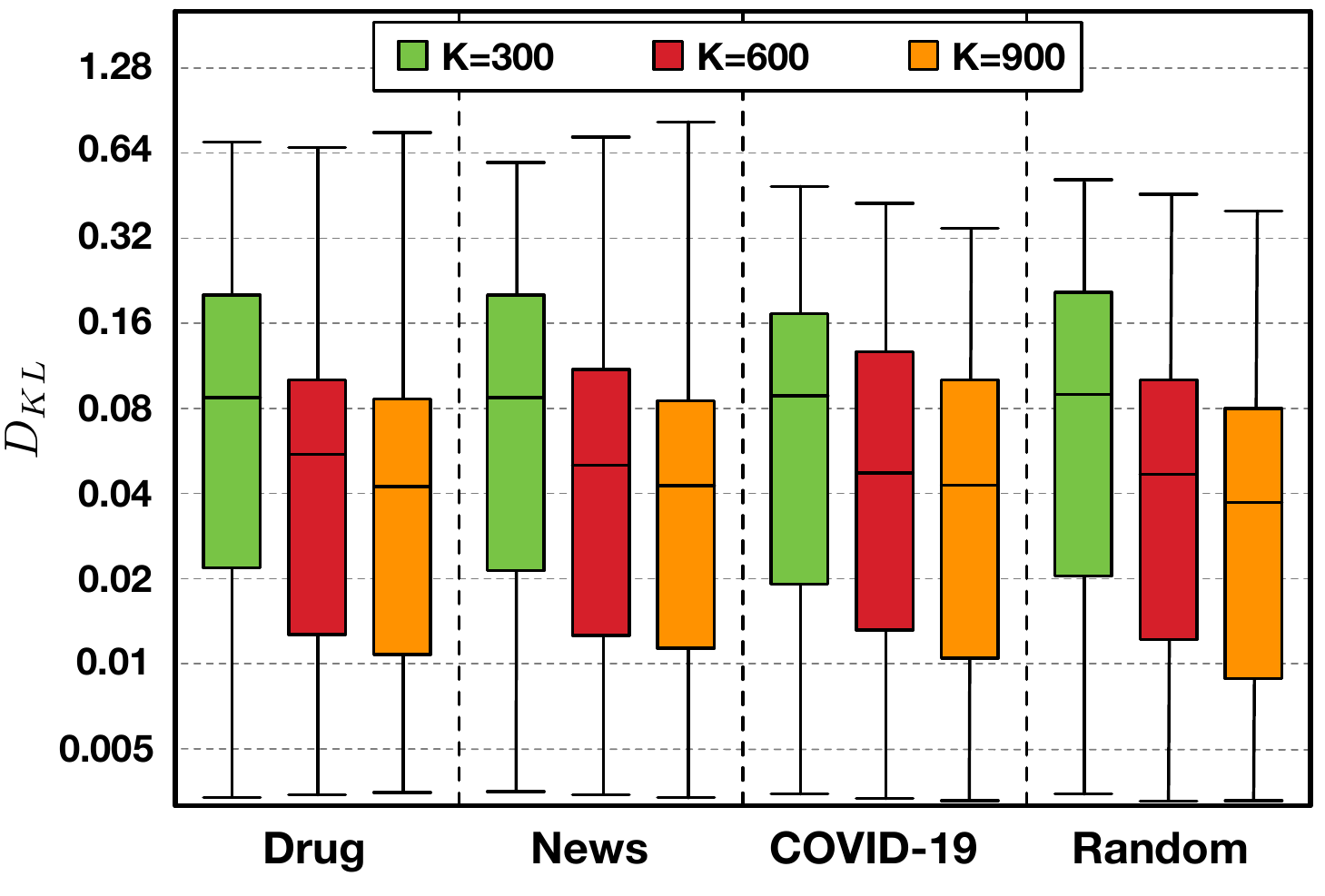}}
   \vspace{-0.2cm}
   \caption{$D_{KL}$ for static arithmetic coding with different $K$s. Note that the $Y$ axis is in the log scale.}
   \label{fig:variable_KL}
   \vspace{-0.3cm}
 \end{figure}

\subsection{Experiment Results}

\smallskip
\noindent \textbf{Overall Performance.}
Table~\ref{table:results} shows the overall performance.
First, we can see all variable length coding algorithms (\ie, RNN-Stega, Patient-Huffman, Arithmetic, \our) outperform the fixed length coding algorithm Bin-LM. 
The Bin-LM method achieves worse imperceptibility (\ie, larger $D_{KL}$ ) when it encodes message bits at higher compression rate (\ie, larger \emph{Bits/Word}), which aligns with the previous theoretical result in~\cite{Dai2019TowardsNS}. 
Second, we observe that Patient-Huffman method improves RNN-Stega as it achieves smaller $D_{KL}$ when \emph{Bits/Word} is kept roughly the same.
Third, we find the arithmetic coding based methods (\ie, Arithmetic and \our) outperform the Huffman tree based methods (\ie, RNN-Stega and Patient-Huffman).
Finally, we can see our proposed \our method can beat Arithmetic by automatically choosing the most appropriate $K$ values and thus achieves the best overall performance.

\begin{figure}[!t]
   \centering
   \centerline{\includegraphics[width=0.48\textwidth]{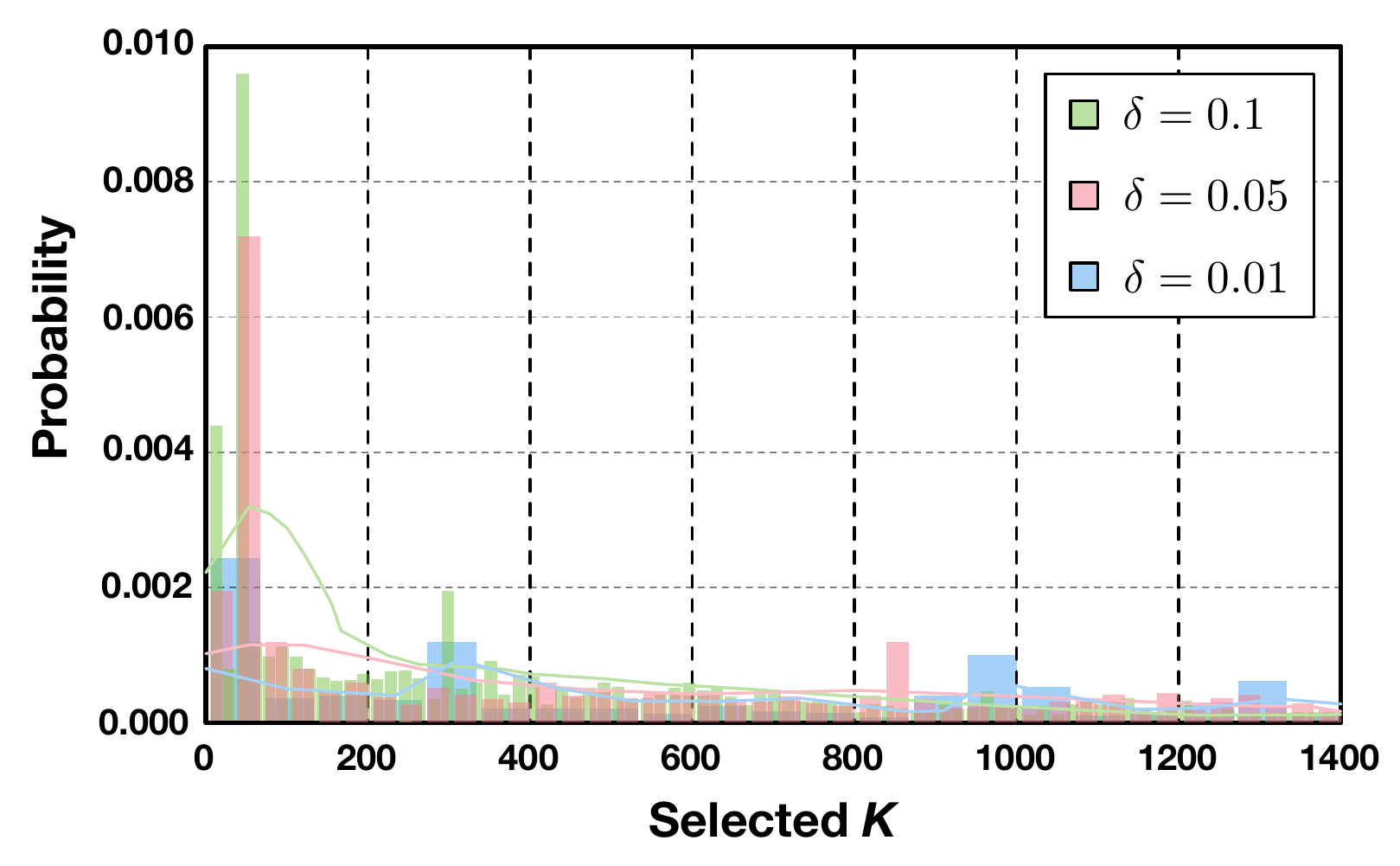}}
   \vspace{-0.2cm}
   \caption{Histogram of selected $K$s in our \our method's next token distribution $\mathbf{Q}(y_t|\mathbf{y}_{<t})$.}
   \label{fig:histK}
   \vspace{-0.3cm}
 \end{figure}

\begin{figure}[!t]
   \centering
   \centerline{\includegraphics[width=0.48\textwidth]{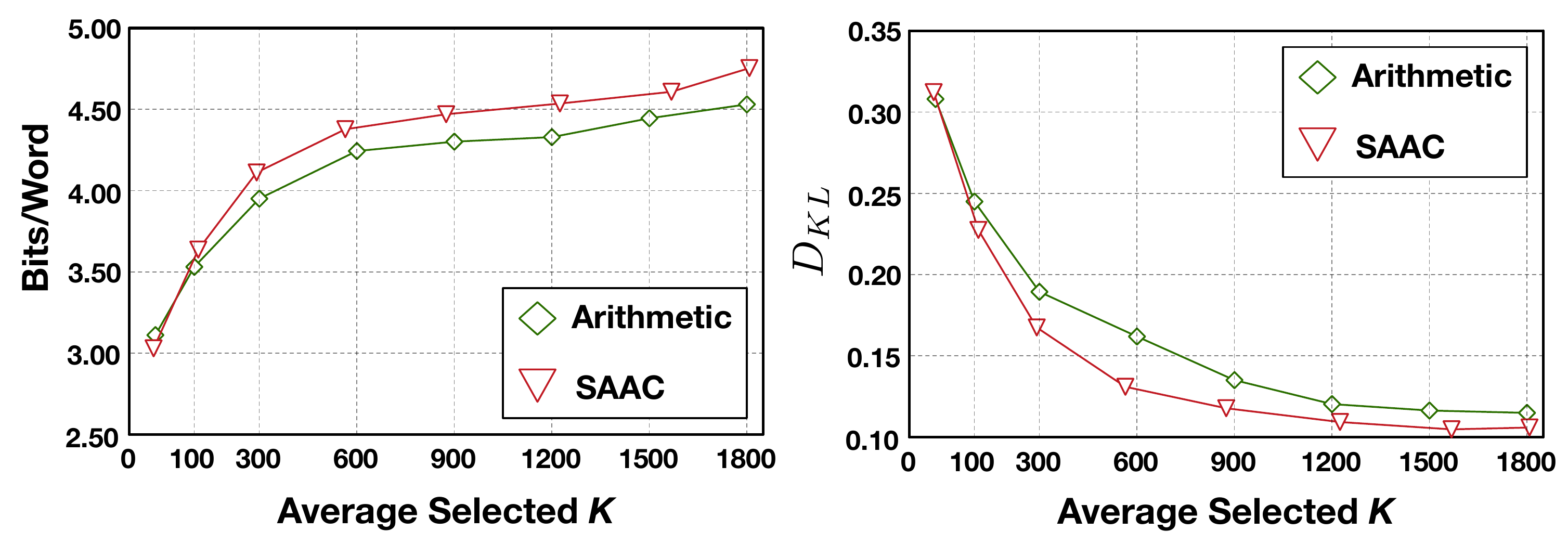}}
   \vspace{-0.2cm}
   \caption{Comparison of baseline Arithmetic method with \our across (roughly) the same average $K$s. Larger \emph{bits/word} values $\uparrow$ and smaller $D_{KL}$ values $\downarrow$ indicate better performance.}
   \label{fig:our_vs_arithmetic}
   \vspace{-0.3cm}
 \end{figure}

\smallskip
\noindent \textbf{Comparison with Arithmetic Baseline.} 
We further analyze where \our's gains over the Arithmetic baseline method come from. 
Fig.~\ref{fig:variable_KL} shows the KL divergence between LM's distribution $\P_{LM}$ and steganography encoder's distribution $\mathbf{Q}$ across all time steps.
We can see that although most of KL values are less than 0.08, the 95th percentiles are all above 0.32, which means even for large predefined $K=900$, five percent of generation steps induce KL values larger than 0.32. 
Fig.~\ref{fig:histK} shows three histograms of \our selected $K$s, one for each required imperceptibility bound $\delta$.
We observe that these histograms have several modes with one (largest) mode around 50 and one mode larger than 300. 
This indicates that for a majority of generation steps, choosing a $K<50$ is enough to guarantee the required imperceptibility bound and thus fixing a static $K=300$ is a big waste for those steps. 
Meanwhile, the LM distributions at some generation steps are too ``flat'' and we indeed need to use a larger $K$ to achieve the required imperceptibility bound $\delta$.
Finally, we vary the imperceptibility bound $\delta$ and calculate the average $K$ selected by \our. 
Fig.~\ref{fig:our_vs_arithmetic} compares the baseline Arithmetic method (of different predefined $K$s) with \our method that has the (roughly) same average selected $K$. 
We can see that using about the same $K$s, our \our method can clearly outperform the Arithmetic baseline method in terms of both \emph{Bits/word} and $D_{KL}$ metrics. 

 \begin{figure*}[!t]
   \centering
   \centerline{\includegraphics[width=0.98\textwidth]{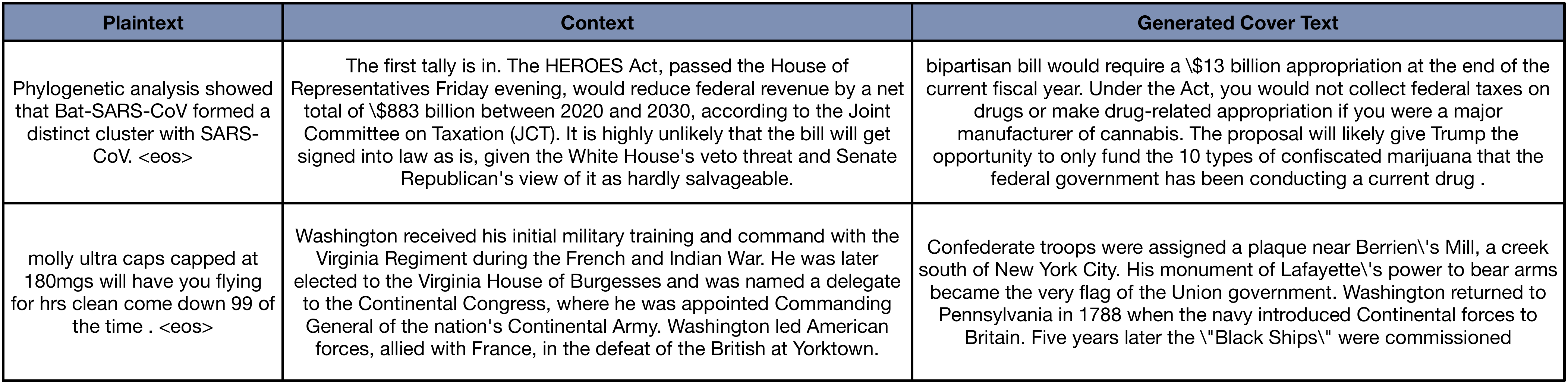}}
    \vspace*{-0.2cm}
    \caption{\label{fig:examples} Cover text examples generated by our \our method. The context is used for generating the first cover text token (c.f. $\mathbf{Q}(\cdot|\mathbf{y}_{<1})$ in Fig.~\ref{fig:arithmetic}). We can see that those generated cover texts are fluent and effectively hide messages in the original plaintexts.}
    \vspace*{-0.3cm}
 \end{figure*}

 \begin{figure*}[!t]
   \centering
   \centerline{\includegraphics[width=0.98\textwidth]{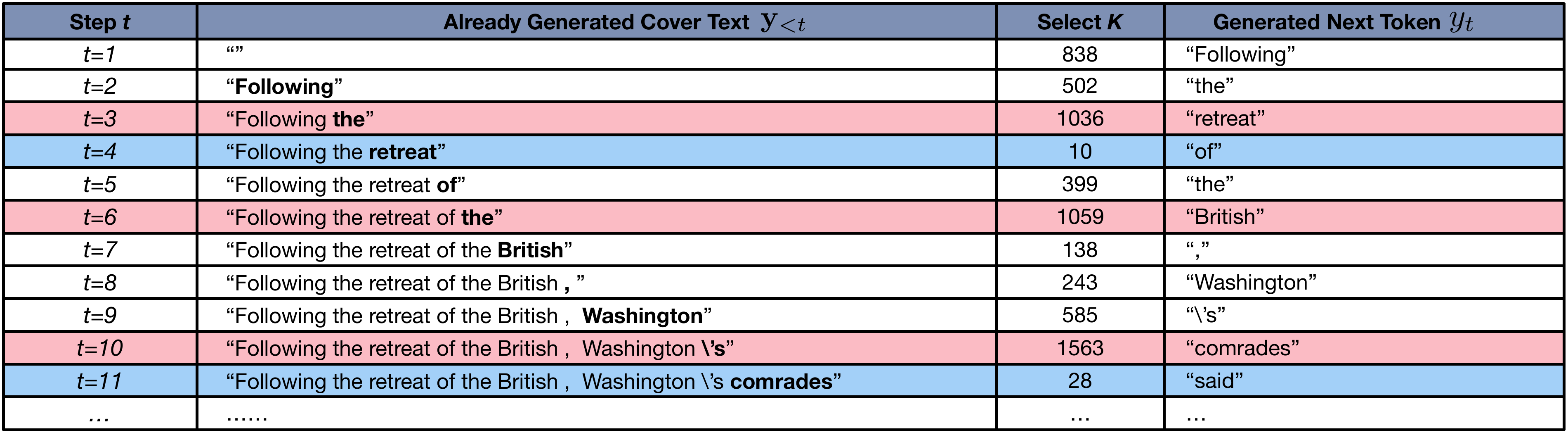}}
    \vspace*{-0.2cm}
    \caption{\label{fig:generation_example} One step-by-step example of cover text generation. When less variety exists in the next token distribution $\mathbf{Q}(\cdot|\mathbf{y}_{<t})$, we will choose a smaller $K$ (lines in blue color). Otherwise, we select a larger $K$ (lines in pink color).}
    \vspace*{-0.3cm}
 \end{figure*}

\smallskip
\noindent \textbf{Efficiency Analysis.}
We run all our experiments on a machine with one single RTX 8000 GPU and 80 Intel Xeon Gold 6230 CPUs.
On average, encoding one sentence takes Bin-LM 2.361 second, RNN-Stega 1.617 second, Arithmetic 2.085 second, Patient-Huffman 4.443 second, and our proposed SAAC method 1.722 second.
This result shows dynamic selection of step-wise $K$ will not introduce many computational overhead and can sometimes even improve the efficiency of the static arithmetic coding method. 

\smallskip
\noindent \textbf{Case Studies.}
We show some concrete examples of generated cover texts in Fig.~\ref{fig:examples}.
Following~\cite{Ziegler2019NeuralLS}, we use an introductory context $c$ for generating the first cover text token (\ie, replace $\mathbf{Q}(\cdot|\mathbf{y}_{<1})$ with $\mathbf{Q}(\cdot|[c; \mathbf{y}_{<1}])$).
This strategy helps to improve the cover text quality and will later also be used in the human evaluation. 
We can see that those generated cover texts are fluent and grammatically correct.
Besides, they are topically similar to the provided introductory context and effectively hide messages in the original plaintexts. 
In Fig.~\ref{fig:generation_example}, we further show a step-by-step generation example. 
We can see that in step 4, the next token distribution $\mathbf{Q}(\cdot|\mathbf{y}_{<4})$ following word ``retreat'' exhibits less variety, and thus we select a small $K=10$.
On the other hand, in step 6, the next token distribution $\mathbf{Q}(\cdot|\mathbf{y}_{<6})$ following word ``the'' has more variety and we use a larger $K=1059$ to satisfy the required imperceptibility.

\subsection{Human Evaluation}

We conduct human evaluation to test whether generated cover texts can indeed fool human eavesdroppers via crowdsourcing.
First, we select 100 news articles from the CNN/DM dataset and treat each article's first 3 sentences as the context.
Next, we sample 100 ciphertexts uniformly at random and pair each of them with the above 3 sentence context. 
Then, for each $\langle$context, ciphertext$\rangle$ pair, we generate a cover text using different steganography methods, including RNN-Stega with Huffman tree depths 3, 5, 7, arithmetic coding with top $K$s 300, 600, 900, and \our with imperceptibility gaps 0.1, 0.05, 0.01. 
Finally, we gather all the generated cover texts; mix them with the original human-written sentences (\ie, the 4th sentence in each news article), and send them to crowd accessors on Amazon Mechanical Turk. 

 \begin{figure}[!t]
   \centering
   \centerline{\includegraphics[width=0.48\textwidth]{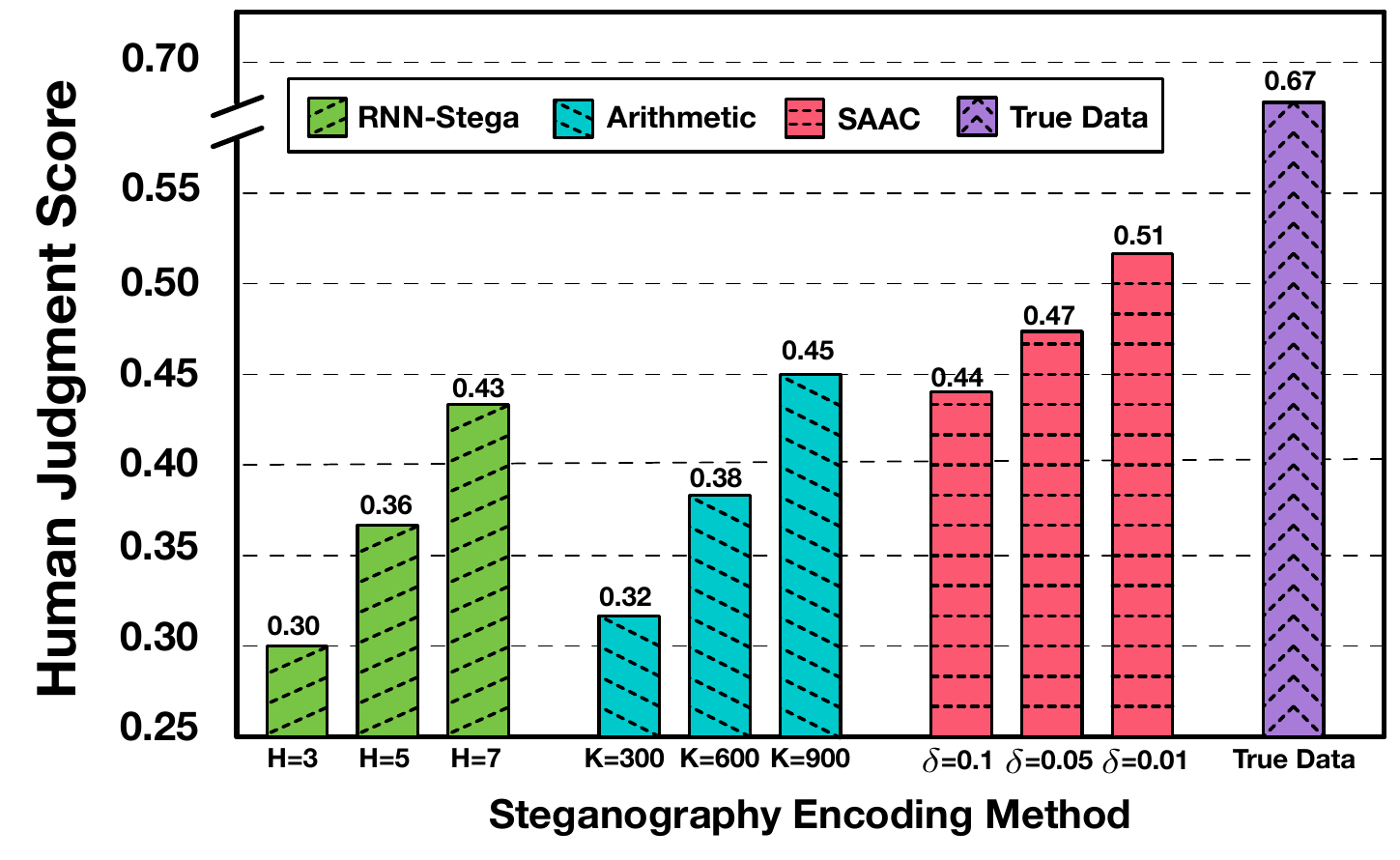}}
   \vspace{-0.2cm}
   \caption{Human evaluation results. $Y$ axis shows the percentage of cover texts (generated by one method) that are considered by humans to be a proper continuation of the context.}
   \label{fig:human}
   \vspace{-0.3cm}
 \end{figure}

In each HIT, the assessor is given one context paired with one sentence and is asked ``Given the start of a news article: <context>, is the following a likely next sentence: <sentence>? Yes or No?''. 
We explicitly ask assessors to consider whether this sentence is grammatically correct, contains no factual error, and makes sense in the given context. 
To ensure the quality of collected data, we require crowd assessors to have a 95\% HIT acceptance rate, a minimum of 1000 HITs, and be located in the United States or Canada. 
Moreover, we include a simple attention check question in 20\% of HITs and filter out the results from assessors who do not pass the attention check.

Fig.~\ref{fig:human} shows the human evaluation results.
First, we can see this test itself is challenging as only 67\% of time people can correctly identify the true follow-up sentence.
Second, more encouragingly, we find the cover texts generated by our \our algorithm can indeed fool humans 51\% of times.
For those cover texts that do not pass the human test, we analyze crowd assessor's feedbacks and find they are rejected mostly because they contain some factual errors.
Thus, we believe improving the generation factual accuracy is an important direction for future linguistic steganography research.

%% file: 5-related_work.tex
\section{Related Work}\label{sec:related_work}

Early steganography methods~\cite{marvel1999spread,gopalan2003audio} use image and audio as the cover signal because they have a high information theoretic entropy.
However, sending an image or audio recording abruptly though a public channel will likely cause the eavesdropper's suspicion. 
Thus, linguistic steganography methods are proposed to leverage text as the cover signal because natural language is prevalent and innocuous in daily life. 

Linguistic steganography methods can be categorized into two types, edit-based or generation-based~\cite{Bennett2004LINGUISTICSS}. 
Edit-based methods try to directly edit the secret message and transform it into an innocent text.
Typical transformations are synonym substitution~\cite{Topkara2006TheHV, Chang2014PracticalLS}, paraphrase substitution~\cite{Chang2010LinguisticSU, Ji2018CreativeLE}, and syntactic transformation~\cite{Thamaraiselvan2015TextbasedSU,Safaka2016MatryoshkaHS}. 
Instead of editing all words in the secret message, \citep{Zhang2014BeAA, Zhang2015ContextawareEM} take an entity-oriented view and focus on encoding/decoding morphs of important entities in the message.
Finally, some work~\cite{grosvald2011free,Wilson2014LinguisticSO} allows human agents to assist the cover text generation process. 

One major limitation of edit-based methods is that they cannot encode information efficiently. 
\citep{Wilson2016AvoidingDO} show the popular CoverTweet system~\cite{Wilson2014LinguisticSO} can encode only two bits information in each transformed tweet on average. 
To address this limitation, generation-based methods try directly output the cover text based on the secret message. 
Early study~\cite{Chapman1997HidingTH} utilizes a generative grammar to output the cover text.
More recently, people leverage a neural language model for linguistic steganography.
One pioneering work by~\citep{Fang2017GeneratingST} divides the message bits into equal-size blocks and encodes each block using one cover text token.
\citep{Yang2019RNNStegaLS} improves the above method by constructing a Huffman tree and encoding the message in variable length chunks via a Huffman tree. 
\citep{Dai2019TowardsNS} presents the first theoretical analysis of the above two methods and proposes a modified Huffman algorithm. 
The method most related to this study is~\cite{Ziegler2019NeuralLS} where the arithmetic coding algorithm is introduced for steganography.
In this study, we present a more formal analysis of arithmetic coding based steganography method and propose a better self-adjusting algorithm to achieve the statistical imperceptibility.

%% file: 6-conclusion.tex
\section{Discussions and Future Work}\label{sec:conclusion}

This work presents a new linguistic steganography method that encodes secret messages using self-adjusting arithmetic coding. 
We formally prove this method is near-imperceptible and empirically show it achieves the state-of-the-art results on various text corpora. 
There are several directions we will further explore in the future.
First, we may combine the edit-based steganography with generative steganography method by first transforming the original plaintext in a semantics-preserving way and then encoding the transformed plaintext. 
Second, we will study whether this current method is still effective when a small-scale neural LM (\eg, distilGPT-2) is applied. 
Finally, this study assumes a passive eavesdropper who does not modify the transmitted cover text.  
Adapting the current methods to be robust to an active eavesdropper who may alter the cover text is another interesting direction.

%% file: 7-ack.tex

\section*{Acknowledgements}
Research was sponsored in part by US DARPA SocialSim Program No. W911NF-17-C-0099, NSF IIS-19-56151, IIS-17-41317, IIS 17-04532, and IIS 16-18481, and DTRA HDTRA11810026. 
Any opinions, findings or recommendations expressed herein are those of the authors and should not be interpreted as necessarily representing the views, either expressed or implied, of DARPA or the U.S. Government. 
We thank anonymous reviewers for valuable and insightful feedback.

%% file: main_stega.bbl
\begin{thebibliography}{33}
\expandafter\ifx\csname natexlab\endcsname\relax\def\natexlab#1{#1}\fi

\bibitem[{Bennett(2004)}]{Bennett2004LINGUISTICSS}
Krista Bennett. 2004.
\newblock Linguistic steganography: Survey, analysis, and robustness concerns
  for hiding information in text.

\bibitem[{Chang and Clark(2010)}]{Chang2010LinguisticSU}
Ching-Yun Chang and Stephen Clark. 2010.
\newblock Linguistic steganography using automatically generated paraphrases.
\newblock In \emph{HLT-NAACL}.

\bibitem[{Chang and Clark(2014)}]{Chang2014PracticalLS}
Ching-Yun Chang and Stephen Clark. 2014.
\newblock Practical linguistic steganography using contextual synonym
  substitution and a novel vertex coding method.
\newblock \emph{Computational Linguistics}, 40:403--448.

\bibitem[{Chapman and Davida(1997)}]{Chapman1997HidingTH}
Mark Chapman and George~I. Davida. 1997.
\newblock Hiding the hidden: A software system for concealing ciphertext as
  innocuous text.
\newblock In \emph{ICICS}.

\bibitem[{Dai and Cai(2019)}]{Dai2019TowardsNS}
Falcon~Z. Dai and Zheng Cai. 2019.
\newblock Towards near-imperceptible steganographic text.
\newblock In \emph{ACL}.

\bibitem[{Duda(2013)}]{Duda2013AsymmetricNS}
Jarek Duda. 2013.
\newblock Asymmetric numeral systems: entropy coding combining speed of huffman
  coding with compression rate of arithmetic coding.

\bibitem[{Fang et~al.(2017)Fang, Jaggi, and Argyraki}]{Fang2017GeneratingST}
Tina Fang, Martin Jaggi, and Katerina~J. Argyraki. 2017.
\newblock Generating steganographic text with lstms.
\newblock In \emph{ACL}.

\bibitem[{Fedotov et~al.(2003)Fedotov, Harremo{\"e}s, and
  Tops{\o}e}]{Fedotov2003RefinementsOP}
Alexei~A. Fedotov, Peter Harremo{\"e}s, and Flemming Tops{\o}e. 2003.
\newblock Refinements of pinsker's inequality.
\newblock \emph{IEEE Trans. Inf. Theory}, 49:1491--1498.

\bibitem[{Gopalan(2003)}]{gopalan2003audio}
Kaliappan Gopalan. 2003.
\newblock Audio steganography using bit modification.
\newblock In \emph{International Conference on Multimedia and Expo}.

\bibitem[{Grosvald and Orgun(2011)}]{grosvald2011free}
Michael Grosvald and C~Orhan Orgun. 2011.
\newblock Free from the cover text: a human-generated natural language approach
  to text-based steganography.
\newblock \emph{Journal of Information Hiding and Multimedia Signal
  Processing}, 2(2):133--141.

\bibitem[{Hermann et~al.(2015)Hermann, Kocisk{\'y}, Grefenstette, Espeholt,
  Kay, Suleyman, and Blunsom}]{Hermann2015TeachingMT}
Karl~Moritz Hermann, Tom{\'a}s Kocisk{\'y}, Edward Grefenstette, Lasse
  Espeholt, Will Kay, Mustafa Suleyman, and Phil Blunsom. 2015.
\newblock Teaching machines to read and comprehend.
\newblock In \emph{NIPS}.

\bibitem[{Huffman(1952)}]{huffman1952method}
David~A Huffman. 1952.
\newblock A method for the construction of minimum-redundancy codes.
\newblock \emph{Proceedings of the IRE}, 40(9):1098--1101.

\bibitem[{Ji and Knight(2018)}]{Ji2018CreativeLE}
Heng Ji and Kevin Knight. 2018.
\newblock Creative language encoding under censorship.
\newblock In \emph{COLING}.

\bibitem[{J{\'o}zefowicz et~al.(2016)J{\'o}zefowicz, Vinyals, Schuster,
  Shazeer, and Wu}]{Jzefowicz2016ExploringTL}
Rafal J{\'o}zefowicz, Oriol Vinyals, Mike Schuster, Noam Shazeer, and Yonghui
  Wu. 2016.
\newblock Exploring the limits of language modeling.
\newblock \emph{ArXiv}.

\bibitem[{Marvel et~al.(1999)Marvel, Boncelet, and Retter}]{marvel1999spread}
Lisa~M Marvel, Charles~G Boncelet, and Charles~T Retter. 1999.
\newblock Spread spectrum image steganography.
\newblock In \emph{IEEE Transactions on image processing}.

\bibitem[{bin Mohamed~Amin et~al.(2003)bin Mohamed~Amin, bt. Salleh, Ibrahim,
  b.~Katmin, and Shamsuddin}]{Amin2003InformationHU}
Muhalim bin Mohamed~Amin, Mazleena bt. Salleh, Subariah Ibrahim, Mohd.~Rozi
  b.~Katmin, and M.~Z.~I. Shamsuddin. 2003.
\newblock Information hiding using steganography.
\newblock \emph{4th National Conference of Telecommunication Technology, 2003.
  NCTT 2003 Proceedings.}, pages 21--25.

\bibitem[{Radford et~al.(2019)Radford, Wu, Child, Luan, Amodei, and
  Sutskever}]{radford2019language}
Alec Radford, Jeff Wu, Rewon Child, David Luan, Dario Amodei, and Ilya
  Sutskever. 2019.
\newblock Language models are unsupervised multitask learners.

\bibitem[{Rissanen and Langdon(1979)}]{Rissanen1979ArithmeticC}
Jorma Rissanen and Glen~G. Langdon. 1979.
\newblock Arithmetic coding.
\newblock \emph{IBM J. Res. Dev.}, 23:149--162.

\bibitem[{Safaka et~al.(2016)Safaka, Fragouli, and
  Argyraki}]{Safaka2016MatryoshkaHS}
Iris Safaka, Christina Fragouli, and Katerina~J. Argyraki. 2016.
\newblock Matryoshka: Hiding secret communication in plain sight.
\newblock In \emph{FOCI}.

\bibitem[{Simmons(1984)}]{simmons1984prisoners}
Gustavus~J Simmons. 1984.
\newblock The prisoners’ problem and the subliminal channel.
\newblock In \emph{Advances in Cryptology}, pages 51--67. Springer.

\bibitem[{Thamaraiselvan and Saradha(2015)}]{Thamaraiselvan2015TextbasedSU}
R.~Thamaraiselvan and A.~Saradha. 2015.
\newblock Text-based steganography using cover text free human-generated
  natural language (hgnl) approach.

\bibitem[{Topkara et~al.(2006)Topkara, Topkara, and Atallah}]{Topkara2006TheHV}
Umut Topkara, Mercan Topkara, and Mikhail~J. Atallah. 2006.
\newblock The hiding virtues of ambiguity: quantifiably resilient watermarking
  of natural language text through synonym substitutions.
\newblock In \emph{Multimedia and Secruity}.

\bibitem[{Tutuncu and Hassan(2015)}]{Tutuncu2015NewAI}
Kemal Tutuncu and Abdikarim~Abi Hassan. 2015.
\newblock New approach in e-mail based text steganography.
\newblock \emph{International Journal of Intelligent Systems and Applications
  in Engineering}, 3:54--56.

\bibitem[{Wang et~al.(2020)Wang, Lo, Chandrasekhar, Reas, Yang, Eide, Funk,
  Kinney, Liu, Merrill, Mooney, Murdick, Rishi, Sheehan, Shen, Stilson, Wade,
  Wang, Wilhelm, Xie, Raymond, Weld, Etzioni, and Kohlmeier}]{Wang2020CORD19TC}
Lucy~Lu Wang, Kyle Lo, Yoganand Chandrasekhar, Russell Reas, Jiangjiang Yang,
  Darrin Eide, Kathryn Funk, Rodney~Michael Kinney, Ziyang Liu, William.
  Merrill, Paul Mooney, Dewey~A. Murdick, Devvret Rishi, Jerry Sheehan, Zhihong
  Shen, Brandon Stilson, Alex~D. Wade, Kuansan Wang, Christopher Wilhelm, Boya
  Xie, Douglas~M. Raymond, Daniel~S. Weld, Oren Etzioni, and Sebastian
  Kohlmeier. 2020.
\newblock Cord-19: The covid-19 open research dataset.
\newblock \emph{ArXiv}, abs/2004.10706.

\bibitem[{Westfeld and Pfitzmann(1999)}]{Westfeld1999AttacksOS}
Andreas Westfeld and Andreas Pfitzmann. 1999.
\newblock Attacks on steganographic systems.
\newblock In \emph{Information Hiding}.

\bibitem[{Wilson et~al.(2014)Wilson, Blunsom, and Ker}]{Wilson2014LinguisticSO}
Alex Wilson, Phil Blunsom, and Andrew~D. Ker. 2014.
\newblock Linguistic steganography on twitter: hierarchical language modeling
  with manual interaction.
\newblock In \emph{Electronic Imaging}.

\bibitem[{Wilson and Ker(2016)}]{Wilson2016AvoidingDO}
Alex Wilson and Andrew~D. Ker. 2016.
\newblock Avoiding detection on twitter: embedding strategies for linguistic
  steganography.
\newblock In \emph{Media Watermarking, Security, and Forensics}.

\bibitem[{Witten et~al.(1987)Witten, Neal, and Cleary}]{Witten1987ArithmeticCF}
Ian~H. Witten, Radford~M. Neal, and John~G. Cleary. 1987.
\newblock Arithmetic coding for data compression.
\newblock \emph{Commun. ACM}, 30:520--540.

\bibitem[{Yang et~al.(2019{\natexlab{a}})Yang, Dai, Yang, Carbonell,
  Salakhutdinov, and Le}]{Yang2019XLNetGA}
Zhilin Yang, Zihang Dai, Yiming Yang, Jaime~G. Carbonell, Ruslan Salakhutdinov,
  and Quoc~V. Le. 2019{\natexlab{a}}.
\newblock Xlnet: Generalized autoregressive pretraining for language
  understanding.
\newblock In \emph{NeurIPS}.

\bibitem[{Yang et~al.(2019{\natexlab{b}})Yang, Guo, Chen, Huang, and
  Zhang}]{Yang2019RNNStegaLS}
Zhong-Liang Yang, Xiaoqing Guo, Zi-Ming Chen, Yongfeng Huang, and Yu-Jin Zhang.
  2019{\natexlab{b}}.
\newblock Rnn-stega: Linguistic steganography based on recurrent neural
  networks.
\newblock \emph{IEEE Transactions on Information Forensics and Security},
  14:1280--1295.

\bibitem[{Zhang et~al.(2014)Zhang, Huang, Pan, Ji, Knight, Wen, Sun, Han, and
  Yener}]{Zhang2014BeAA}
Boliang Zhang, Hongzhao Huang, Xiaoman Pan, Heng Ji, Kevin Knight, Zhen Wen,
  Yizhou Sun, Jiawei Han, and B{\"u}lent Yener. 2014.
\newblock Be appropriate and funny: Automatic entity morph encoding.
\newblock In \emph{ACL}.

\bibitem[{Zhang et~al.(2015)Zhang, Huang, Pan, Li, Lin, Ji, Knight, Wen, Sun,
  Han, and Yener}]{Zhang2015ContextawareEM}
Boliang Zhang, Hongzhao Huang, Xiaoman Pan, Sujian Li, Chin-Yew Lin, Heng Ji,
  Kevin Knight, Zhen Wen, Yizhou Sun, Jiawei Han, and B{\"u}lent Yener. 2015.
\newblock Context-aware entity morph decoding.
\newblock In \emph{ACL}.

\bibitem[{Ziegler et~al.(2019)Ziegler, Deng, and Rush}]{Ziegler2019NeuralLS}
Zachary~M. Ziegler, Yuntian Deng, and Alexander~M. Rush. 2019.
\newblock Neural linguistic steganography.
\newblock In \emph{EMNLP}.

\end{thebibliography}
